\documentclass{article}
\usepackage{spconf,amsmath,graphicx,subfigure}
\usepackage{tabularx}
\usepackage{color}
\usepackage{dcolumn}
\usepackage{subfig}
\usepackage{tabularx}

\usepackage{setspace}
\title{Generative Adversarial Networks for Image-to-Image Translation on Multi-Contrast MR Images - A Comparison of CycleGAN and UNIT}
\name{Per Welander$^{\: a}$ Simon Karlsson$^{\: a}$ Anders Eklund$^{\: a,b,c}$}
\address{$^a$Division of Medical Informatics, Department of Biomedical Engineering, \\ Link\"{o}ping University, Link\"{o}ping, Sweden \\ $^b$Division of Statistics and Machine Learning, Department of Computer and Information Science, \\ Link\"{o}ping University, Link\"{o}ping, Sweden \\ $^c$Center for Medical Image Science and Visualization (CMIV), \\ Link\"{o}ping University, Link\"{o}ping, Sweden}

\begin{document}
%\ninept
%
\maketitle

\thispagestyle{plain}
\pagestyle{plain}

\begin{abstract}

In medical imaging, a general problem is that it is costly and time consuming to collect high quality data from healthy and diseased subjects. Generative adversarial networks (GANs) is a deep learning method that has been developed for synthesizing data. GANs can thereby be used to generate more realistic training data, to improve classification performance of machine learning algorithms. Another application of GANs is image-to-image translations, e.g. generating magnetic resonance (MR) images from computed tomography (CT) images, which can be used to obtain multimodal datasets from a single modality. Here, we evaluate two unsupervised GAN models (CycleGAN and UNIT) for image-to-image translation of T1- and T2-weighted MR images, by comparing generated synthetic MR images to ground truth images. We also evaluate two supervised models; a modification of CycleGAN and a pure generator model. A small perceptual study was also performed to evaluate how visually realistic the synthesized images are. It is shown that the implemented GAN models can synthesize visually realistic MR images (incorrectly labeled as real by a human). It is also shown that models producing more visually realistic synthetic images not necessarily have better quantitative error measurements, when compared to ground truth data. Code is available at https://github.com/simontomaskarlsson/GAN-MRI.

\end{abstract}

\section{Introduction}
Deep learning has been applied in many different research fields to solve complicated problems~\cite{LeCun2015}, made possible through parallel computing and big datasets. Acquiring a large annotated medical imaging dataset can be rather challenging for classification problems (e.g. discriminating healthy and diseased subjects), as one training example then corresponds to one subject~\cite{Litjens}. Data augmentation, e.g. rotation, cropping and scaling, is normally used to increase the amount of training data, but can only provide limited alternative data. A more advanced data augmentation technique, generative adversarial networks (GANs)~\cite{Goodfellow2014}, uses two competing convolutional neural networks (CNNs); one that generates new samples from noise and one that that discriminates samples as real or synthetic. The most obvious application of a GAN in medical imaging is to generate additional realistic training data, to improve classification performance (see e.g. ~\cite{Antoniou2017} and~\cite{Calimeri2017}). Another application is to use GANs for image-to-image translation, e.g. to generate computed tomography (CT) data from magnetic resonance (MR) images or vice versa. This can for example be very useful for multimodal classification of healthy and diseased subjects, where several types of medical images (e.g. CT and MRI) are combined to improve sensitivity (see e.g.~\cite{Dai2012} and~\cite{Zhang2011}). 

To use GANs for image-to-image translation in medical imaging is not a new idea. Nie et al.~\cite{Nie2016} used a GAN to generate CT data from MRI. Yang et al.~\cite{Yang2018} recently used GANs to improve registration and segmentation of MR images, by generating new data and using multimodal algorithms. Similarly, Dar et al.~ \cite{Hassan2018} demonstrate how GANs can be used for generation of a T2-weighted MR image from a T1-weighted image. However, since GANs have only recently been proposed for image-to-image translation, and new GAN models are still being developed, it is not clear what the best GAN model is and how GANs should be evaluated and compared. We therefore present a small comparison for image-to-image translation of T1- and T2-weighted MR images. Compared to previous work~\cite{Yang2018,Hassan2018} which used conditional GANs (cGAN)~\cite{isola2016image}, we show results for our own Keras implementations of CycleGAN~\cite{cycleGANs} and UNIT~\cite{VAECoupledGANs}, see https://github.com/simontomaskarlsson/GAN-MRI for code.

\section{Method}

\subsection{GAN model selection and implementation}

Several different GAN models were investigated in a literature study~\cite{cycleGANs,VAECoupledGANs,StarGANs,GeneGANs,DualGANs}. Two models stood out among the others in synthesizing realistic images in high resolution; CycleGAN~\cite{cycleGANs} and UNIT~\cite{VAECoupledGANs}. Training of neural networks is commonly supervised, i.e. the training requires corresponding ground truth to each input sample. In image-to-image translation this means that paired images from both source and target domain are needed. To alleviate this constraint, CycleGAN and UNIT can work with unpaired training data.

Two different variants of the CycleGAN model were implemented, \emph{CycleGAN\_s} and \emph{CycleGAN}. Including the ground truth image in the training should intuitively generate better results, since the model then has more information about how the generated image should appear. To investigate this, \emph{CycleGAN\_s} was implemented and trained supervised, i.e. by adding the mean absolute error (MAE) between output and ground truth data. To investigate how the adversarial and cyclic loss contribute to the model, \emph{Generators\_s} was also implemented. It consists of the generators in \emph{CycleGAN} and is only trained in a supervised manner with a MAE loss using the ground truth images, it does not include the adversarial or the cyclic loss. A \emph{Simple} baseline model was also implemented for comparison, it consists of only two convolutional layers.

\subsection{Evaluation}

The dataset used in the evaluation is provided by the Human Connectome project~\cite{van2013wu,glasser2013} (https://ida.loni.usc.edu/login.jsp)\footnote{Data collection and sharing for this project was provided by the Human Connectome Project (U01-MH93765) (HCP; Principal Investigators: Bruce Rosen, M.D., Ph.D., Arthur W. Toga, Ph.D., Van J.Weeden, MD). HCP funding was provided by the National Institute of Dental and Craniofacial Research (NIDCR), the National Institute of Mental Health (NIMH), and the National Institute of Neurological Disorders and Stroke (NINDS). HCP data are disseminated by the Laboratory of Neuro Imaging at the University of Southern California.}. We used (paired) T1- and T2-weighted images from 1113 subjects (but note that CycleGAN and UNIT can be trained using unpaired data). All the images have been registered to a common template brain, such that they are in the same position and of the same size. We used axial images (only slice 120) from the segmented brains. The data were split into a training set of 900 images in each domain. The remaining 213 images, in each domain, were used for testing. Using an Nvidia 12 GB Titan X GPU, training times for CycleGAN and UNIT were 419 and 877 seconds/epoch, respectively. The models were on average trained using 180 epochs. The generation of synthetic images took 0.0176 and 0.0478 ms per image for CycleGAN and UNIT, respectively.

The two GAN models were compared using quantitative and qualitative methods. All quantitative results; MAE, mutual information (MI), and peak signal to noise ratio (PSNR), are based on the test dataset. Since the MR images can naturally differ in intensity, each image is normalized before the calculations by division of the standard deviation and subtraction of the mean value.

To visually evaluate a synthetic image compared to a real image can be difficult if the differences are small. A solution to the visual inspection is to instead visualize a relative error between the real image and the synthetic image. This is done by calculating the absolute difference between the images, and dividing it by the real image. These calculations are done on images normalized in the same manner as for the quantitative evaluation, and the error is the relative absolute difference.

Determining if the synthetic MR images are visually realistic or not was done via a perceptual study by one of the authors (Anders Eklund). The evaluator received T1- and T2-weighted images where 96 of them were real and 72 were synthetic, the evaluator then had to determine if each image was real or synthetic. The real and synthetic images were equally divided between the two domains. Images from \emph{Generators\_s} and \emph{Simple} were not evaluated since it is obvious that the images are synthetic, due to the high smoothness. Evaluating anatomical images is a complicated task best performed by a radiologist. The results presented in this paper should therefore only be seen as an indicator of the visual quality.

\section{Results}
\begin{figure*}[]
    \newcommand{\scaling}{0.3}
    \centering
    \begin{tabular}{c c}
         \includegraphics[scale=\scaling]{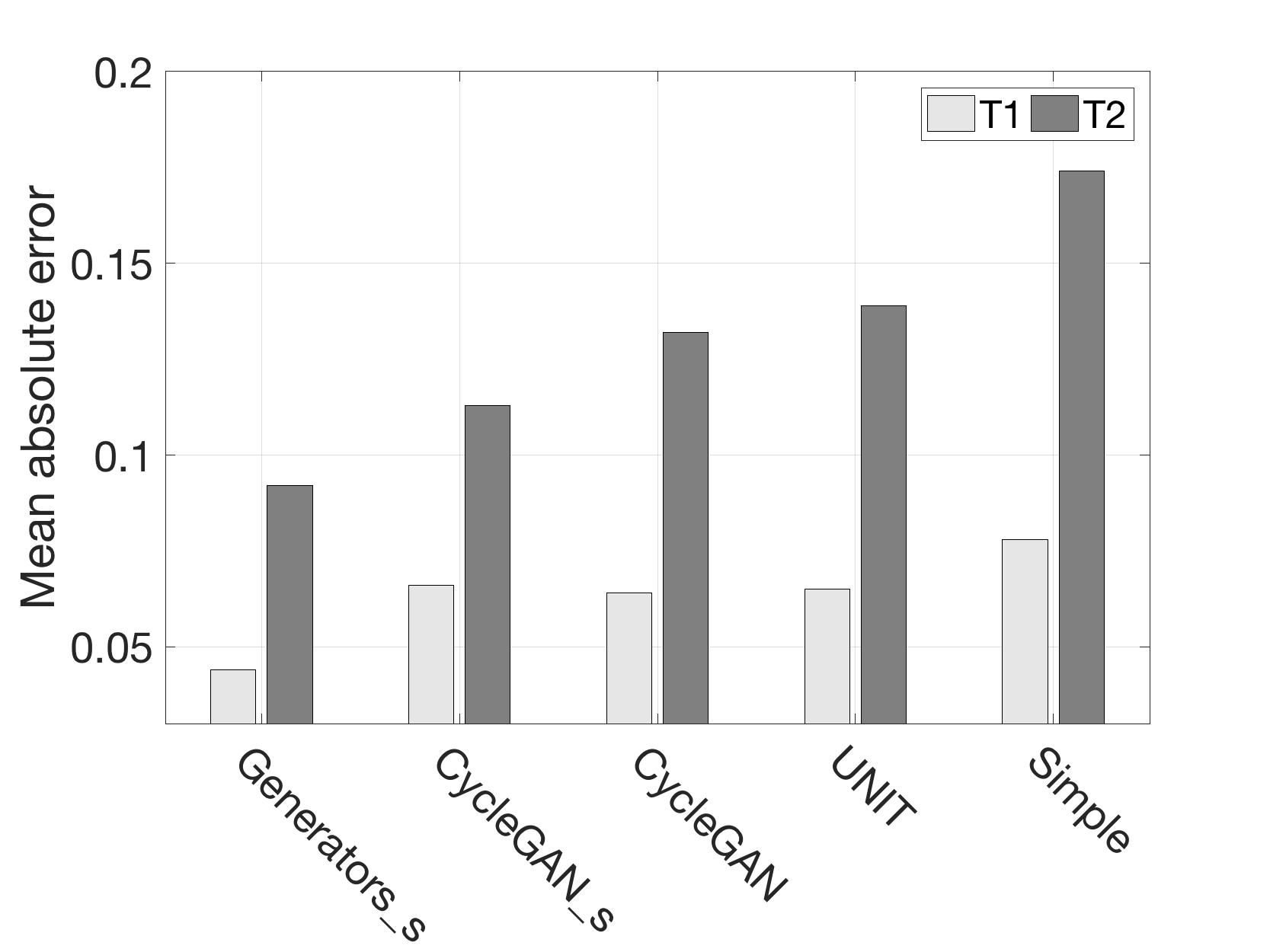} & \includegraphics[scale=\scaling]{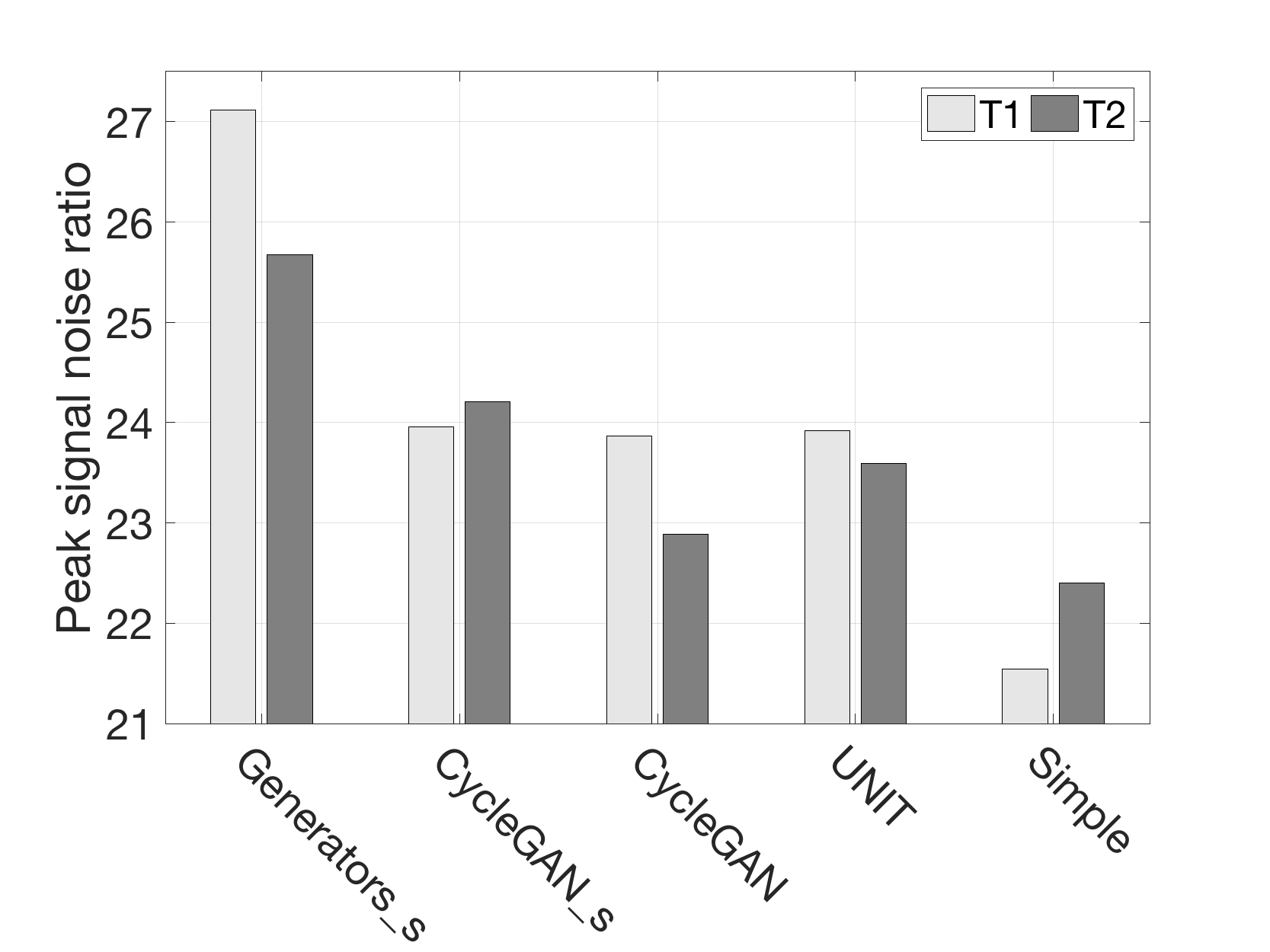} \\
         (a) & (b) \\ \vspace{3mm}
    \end{tabular}
        \includegraphics[scale=\scaling]{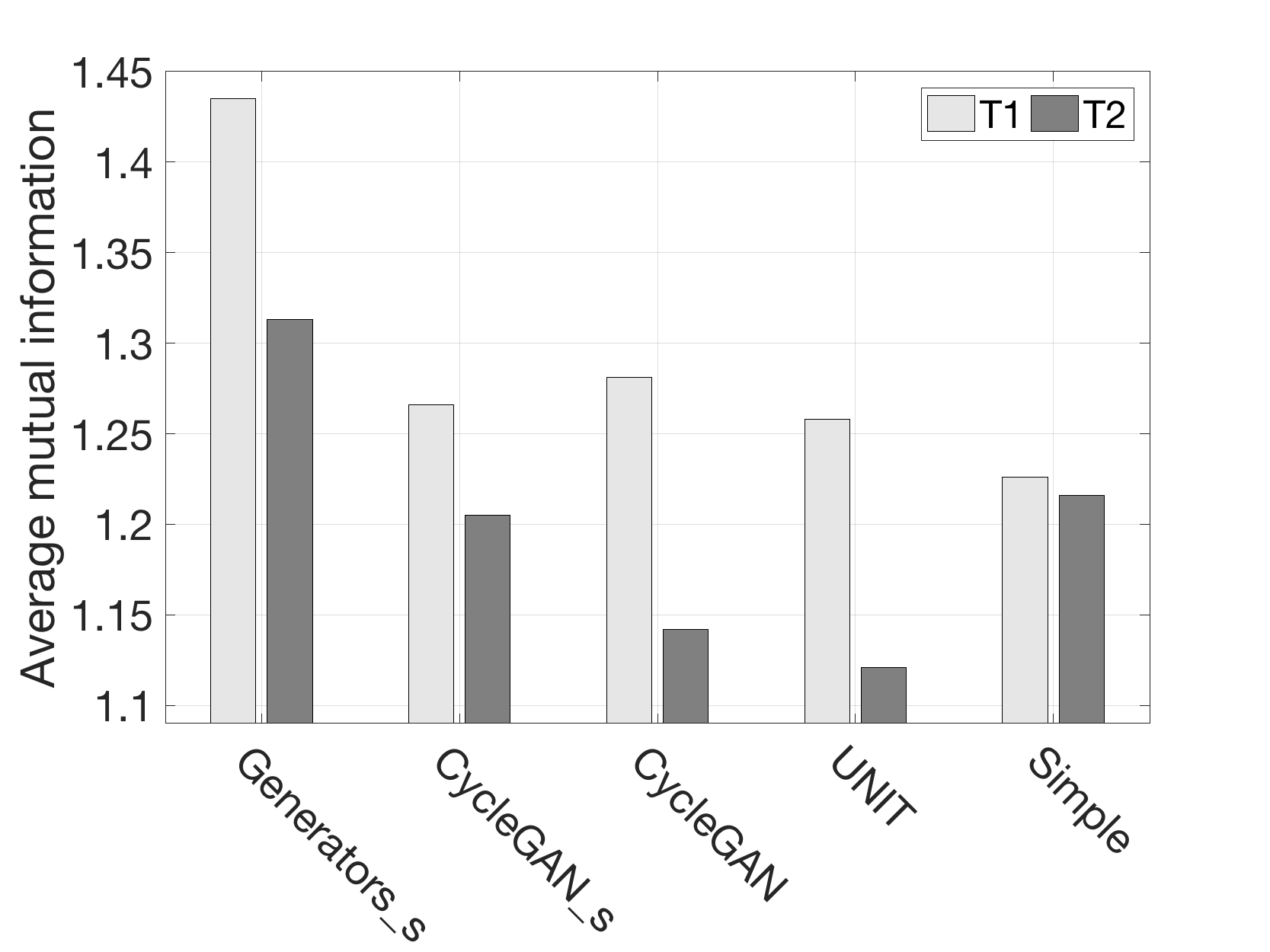} \\
        (c) \\ \vspace{7mm}
        
    \begin{tabular}{c c}     
        \includegraphics[scale=\scaling]{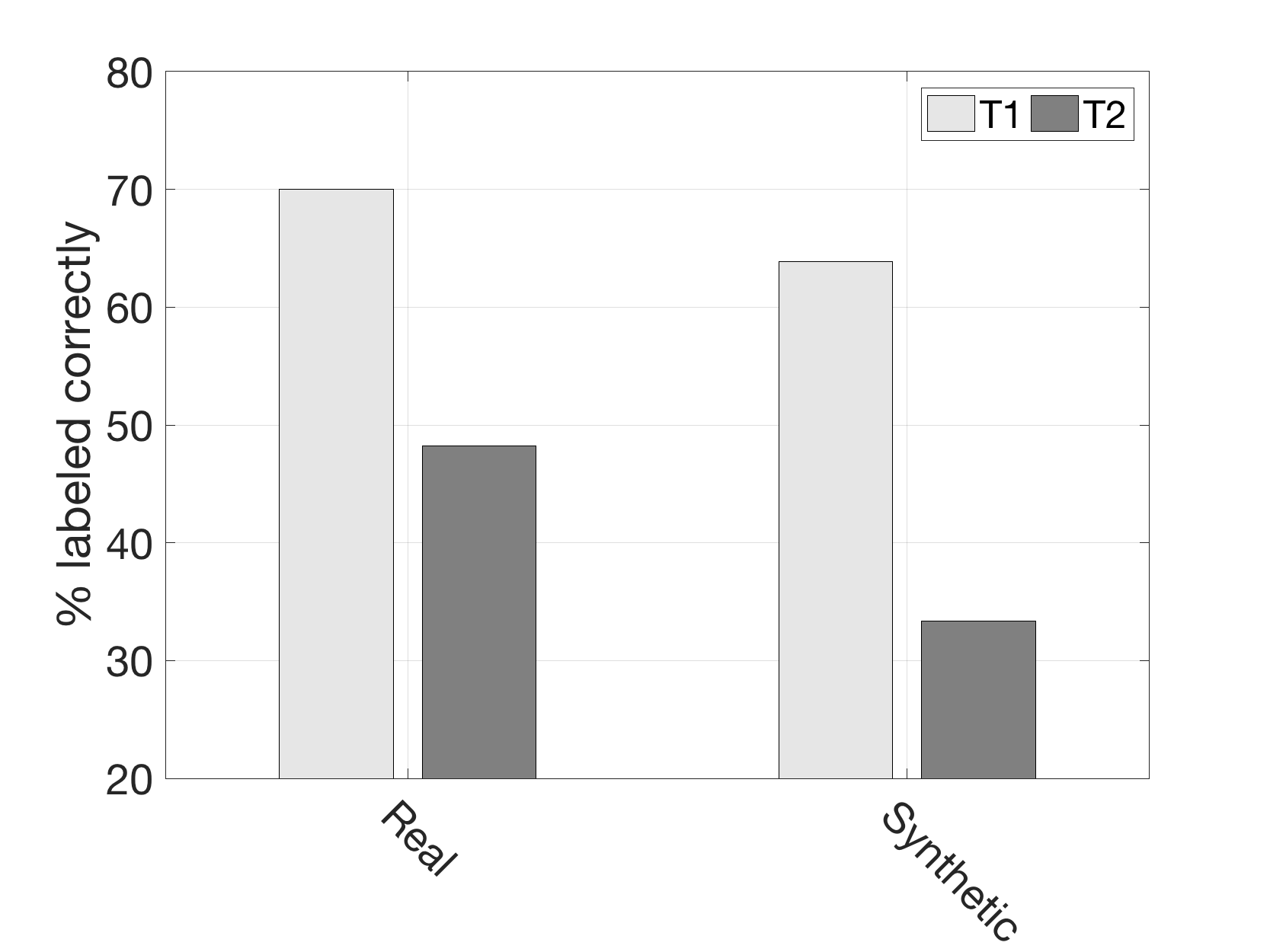}  & \includegraphics[scale=\scaling]{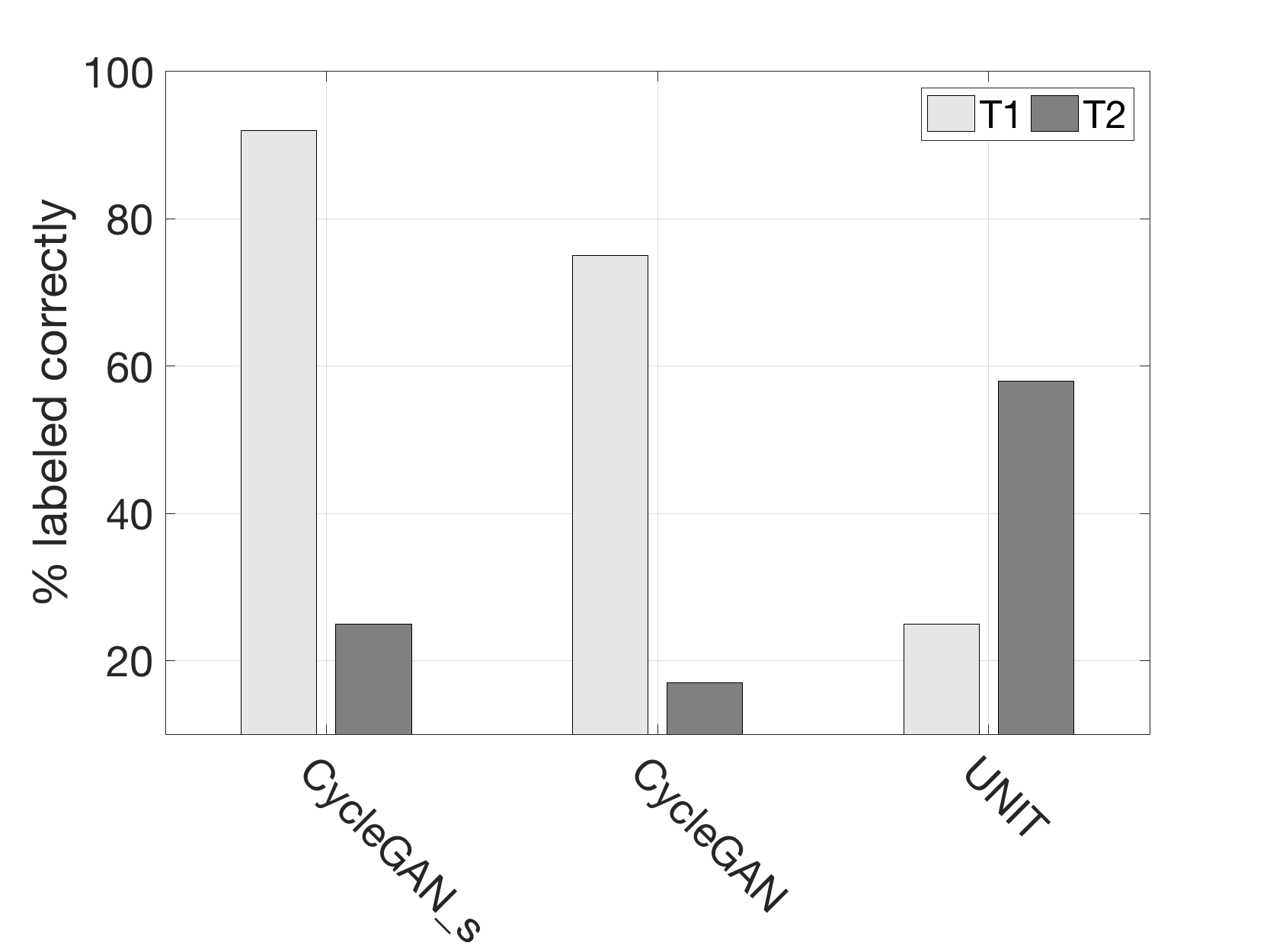} \\
         (d) & (e) \\
    \end{tabular}
    \caption{Quantitative error measurements: \textbf{(a)} - MAE, \textbf{(b)} - PSNR and \textbf{(c)} - MI, for the compared GAN models. The results in \textbf{(d)} are the total scores of all GAN models in the perceptual study, and the results in \textbf{(e)} are for each specific model. Labeling T2-weighted images as real or synthetic is harder due to the fact that T2 images are darker by nature.
    }
    \label{fig:1}
\end{figure*}

Quantitative results and results from the perceptual study are shown in Figure~\ref{fig:1}. The \emph{Generators\_s} model outperforms the other models in all quantitative measurements. The worst performing model on all quantitative measurements, besides MI on T2 images, is the \emph{Simple} model (despite its supervised nature equivalent to \emph{Generators\_s}). The performance of \emph{CycleGAN}, \emph{CycleGAN\_s} and \emph{UNIT} is similar. With just a few exceptions, the quantitative performance is better for T1 images. The opposite is however shown in the perceptual study where more synthetic T1 images are labeled as synthetic compared to T2. Opposite results are in the perceptual study attained for \emph{CycleGAN} and \emph{UNIT}, where \emph{UNIT} shows the best performance for T1 images and \emph{CycleGAN} shows the best performance for T2 images.

\begin{figure*}
    \centering
    \includegraphics[width=0.9\textwidth]{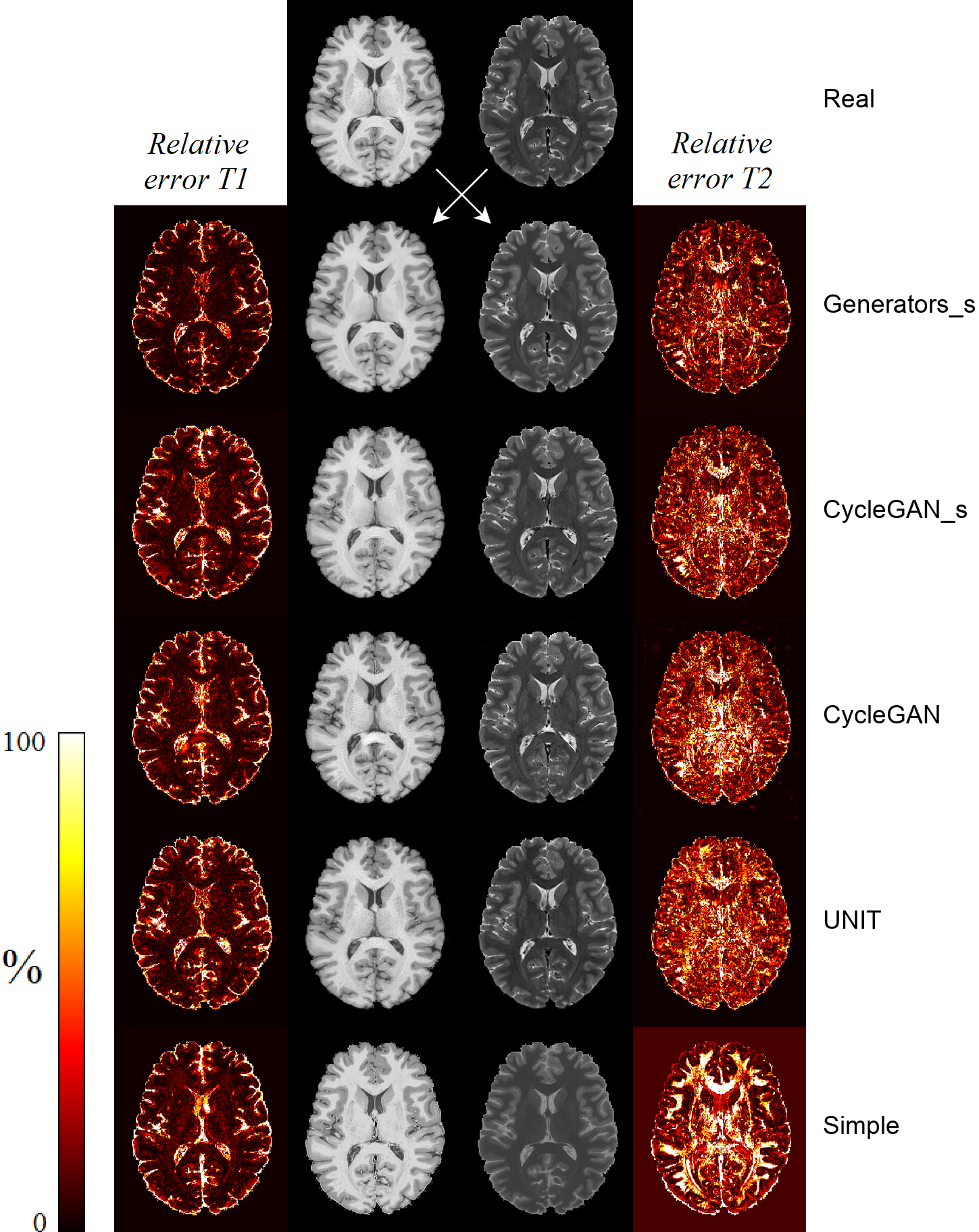}
    \caption{Synthetic images from the evaluated GAN models. The real images shown at the top are inputs that the synthetic images are based on, this is clarified by the white arrows. The real T2 image is the input that generated the synthetic T1 images and vice versa. The images are the same slice from a single subject. This means that the top images are ground truth for the images below them. The colorbar belongs to the images in the left and right columns, which are calculated as the relative absolute difference between the synthetic and the ground truth image. T1 results are shown in the far left column, and T2 results in the far right column.}
    \label{fig:2}
\end{figure*}

The quantitative superiority of \emph{Generators\_s} does not correspond to the visual realness shown in Figure~\ref{fig:2}. The supervised training results in an unrealistic, smooth appearance seen in the MR images from \emph{Generators\_s} and the \emph{Simple} model, where the \emph{Simple} model also fails in the color mapping of the cerebrospinal fluid. The GAN models trained using an adversarial loss generate more realistic synthetic MR images.

The relative absolute error images in Figure~\ref{fig:2} show a greater error for the synthetic T2 images compared to the synthetic T1 images. Synthetic T1 images especially have problems at the edges, whereas errors in the T2 images appear all over the brain.

\section{Discussion}

\subsection{Quantitative comparison}

During training the \emph{Generators\_s} model uses MAE as its only loss function, which creates a model where the goal is to minimize the MAE. The model does this well compared to other models, as shown in Figure~\ref{fig:1}. The \emph{Simple} model, which similarly to \emph{Generators\_s} is only trained using the MAE loss, has the highest error among the models. The \emph{Simple} model only has two convolutional layers and \emph{Generators\_s} has, similar to the CycleGAN generators, 24 convolutional layers. This indicates that the architecture in the \emph{Simple} model is not sufficiently complex for the translation. 

As expected, the \emph{CycleGAN\_s} model shows a slight improvement in MAE for T2 images compared to \emph{CycleGAN}. However, the results are not significantly better than \emph{CycleGAN} and the MAE on T1 images is in fact better for \emph{CycleGAN}. The \emph{CycleGAN} and \emph{UNIT} show similar results and it is difficult to argue why one or the other performs slightly better than the other one.

Figure~\ref{fig:1}c shows that T1 images have a higher MI value than T2 images. This can be correlated to the results from MAE where a larger error was generated from the T2 images. An explanation to why the \emph{Simple} model has a higher MI score than the majority of models for T2 images, is that T1 and T2 images from the same subject contain very similar information. Since the \emph{Simple} model only changes the pixel intensity, the main information is preserved.

\subsection{Qualitative comparison}

From the perceptual study it was shown that the synthetic images have a visually realistic appearance, since synthetic images were classified as real. T2 images were more difficult to classify than T1 images and the reason for the difference can be that the synthetic T2 images had a more realistic appearance, but also the darker nature of T2 images (which for example makes it more difficult to determine if the noise is realistic or not).

The large error on the edges of the synthetic brain images in Figure~\ref{fig:2} can be explained by the fact that each brain has a unique shape, and that T2 images are bright for CSF. Areas where there is an intensity change, e.g. CSF and white matter, seem to be more difficult for the models to learn, this might also be due to differences between subjects.

The \emph{CycleGAN\_s} penalizes appearance different from the ground truth, since it uses the MAE loss during training, which forces it to another direction, closer to the smooth appearance of the images from the \emph{Generators\_s} model. If the aim of the test would instead be to evaluate how similar the synthetic images are to the ground truth, the translated images from \emph{CycleGAN\_s} may give better results.

From the results in Figure~\ref{fig:2} it is obvious that the supervised training, using MAE, pushes the generators into producing smooth synthetic brain images. Another loss function would probably alter the results, but since it is difficult to create mathematical expressions for assessing how realistic an image is, obtaining visually realistic results using supervised methods is a problematic task. The adversarial loss created by the GAN framework allows the discriminator to act as the complex expression, which results in visually realistic images created from the GAN models.

If the aim was to create images that are as similar to ground truth images as possible, the quantitative measurements would be more applicable. It is clear that even if a model such as \emph{Simple} has a relatively good score in the quantitative measurements, it does not necessarily generate visually realistic images. This indicates that solely determining if an image is visually realistic can not be done with the used metrics.

\subsection{Future work}

It has been shown, via a perceptual study, that CycleGAN and UNIT can be used to generate visually realistic MR images. The models performed differently in generating images in the different domains, and training CycleGAN in an unsupervised manner is a better alternative if the aim is to generate as visually realistic images as possible.

A suggestion for future work is to investigate if GANs can be used for data augmentation (e.g. for discriminating healthy and diseased subjects). This would also provide information regarding if the model which creates the most visually realistic images, or the model which performs best in the quantitative evaluations, is the most suitable to use. Here we have only used 2D GANs, but 3D GANs~\cite{Nie2016,Yu2018} can potentially yield even better results, at the cost of a longer processing time and an increased memory usage.

\section{Acknowledgments}

This study was supported by Swedish research council grant 2017-04889. Funding was also provided by the Center for Industrial Information Technology (CENIIT) at Link\"{o}ping University, and the Knut and Alice Wallenberg foundation project "Seeing organ function".

\clearpage
\newpage
\bibliographystyle{IEEEbib}
\bibliography{test}

\begin{thebibliography}{10}

\bibitem{LeCun2015}
Yann LeCun, Yoshua Bengio, and Geoffrey Hinton,
\newblock ``Deep learning,''
\newblock {\em Nature}, vol. 521, pp. 436--444, 2015.

\bibitem{Litjens}
Geert Litjens, Thijs Kooi, Babak~Ehteshami Bejnordi, Arnaud Arindra~Adiyoso
  Setio, Francesco Ciompi, Mohsen Ghafoorian, Jeroen~A.W.M. van~der Laak, Bram
  van Ginneken, and Clara~I. Sanchez,
\newblock ``A survey on deep learning in medical image analysis,''
\newblock {\em Medical Image Analysis}, vol. 42, pp. 60 -- 88, 2017.

\bibitem{Goodfellow2014}
Ian Goodfellow, Jean Pouget-Abadie, Mehdi Mirza, Bing Xu, David Warde-Farley,
  Sherjil Ozair, Aaron Courville, and Yoshua Bengio,
\newblock ``Generative adversarial nets,''
\newblock in {\em Advances in Neural Information Processing Systems 27}, pp.
  2672--2680. 2014.

\bibitem{Antoniou2017}
Antreas Antoniou, Amos Storkey, and Harrison Edwards,
\newblock ``Data augmentation generative adversarial networks,''
\newblock {\em arXiv}, vol. 1711.04340, 2017.

\bibitem{Calimeri2017}
Francesco Calimeri, Aldo Marzullo, Claudio Stamile, and Giorgio Terracina,
\newblock ``Biomedical data augmentation using generative adversarial neural
  networks,''
\newblock in {\em {Artificial Neural Networks and Machine Learning (ICANN)}},
  Alessandra Lintas, Stefano Rovetta, Paul~F.M.J. Verschure, and
  Alessandro~E.P. Villa, Eds., 2017, pp. 626--634.

\bibitem{Dai2012}
Dai Dai, Jieqiong Wang, Jing Hua, and Huiguang He,
\newblock ``Classification of {ADHD} children through multimodal magnetic
  resonance imaging,''
\newblock {\em Frontiers in Systems Neuroscience}, vol. 6, pp. 63, 2012.

\bibitem{Zhang2011}
Daoqiang Zhang, Yaping Wang, Luping Zhou, Hong Yuan, and Dinggang Shen,
\newblock ``Multimodal classification of {Alzheimer's} disease and mild
  cognitive impairment,''
\newblock {\em NeuroImage}, vol. 55, no. 3, pp. 856 -- 867, 2011.

\bibitem{Nie2016}
Dong Nie, Roger Trullo, Caroline Petitjean, Su~Ruan, and Dinggang Shen,
\newblock ``{Medical Image Synthesis with Context-Aware Generative Adversarial
  Networks},''
\newblock {\em arXiv}, vol. 1612.05362, 2016.

\bibitem{Yang2018}
Qianye Yang, Nannan Li, Zixu Zhao, Xingyu Fan, Eric I-Chao Chang, and Yan Xu,
\newblock ``{MRI Image-to-Image Translation for Cross-Modality Image
  Registration and Segmentation},''
\newblock {\em arXiv}, vol. 1801.06940, 2018.

\bibitem{Hassan2018}
Salman Ul~Hassan Dar, Mahmut Yurt, Levent Karacan, Aykut Erdem, Erkut Erdem,
  and Tolga Çukur,
\newblock ``{Image Synthesis in Multi-Contrast MRI with Conditional Generative
  Adversarial Networks},''
\newblock {\em arXiv}, vol. 1802.01221, 2018.

\bibitem{isola2016image}
P.~{Isola}, J.-Y. {Zhu}, T.~{Zhou}, and A.~A. {Efros},
\newblock ``{Image-to-Image Translation with Conditional Adversarial
  Networks},''
\newblock {\em ArXiv:1611.07004}, Nov. 2016.

\bibitem{cycleGANs}
J.-Y. {Zhu}, T.~{Park}, P.~{Isola}, and A.~A. {Efros},
\newblock ``{Unpaired Image-to-Image Translation using Cycle-Consistent
  Adversarial Networks},''
\newblock {\em ArXiv:1703.10593}, Mar. 2017.

\bibitem{VAECoupledGANs}
M.-Y. {Liu}, T.~{Breuel}, and J.~{Kautz},
\newblock ``{Unsupervised Image-to-Image Translation Networks},''
\newblock {\em ArXiv:1703.00848}, Mar. 2017.

\bibitem{StarGANs}
Y.~{Choi}, M.~{Choi}, M.~{Kim}, J.-W. {Ha}, S.~{Kim}, and J.~{Choo},
\newblock ``{StarGAN: Unified Generative Adversarial Networks for Multi-Domain
  Image-to-Image Translation},''
\newblock {\em ArXiv:1711.09020}, Nov. 2017.

\bibitem{GeneGANs}
S.~{Zhou}, T.~{Xiao}, Y.~{Yang}, D.~{Feng}, Q.~{He}, and W.~{He},
\newblock ``{GeneGAN: Learning Object Transfiguration and Attribute Subspace
  from Unpaired Data},''
\newblock {\em ArXiv:1705.04932}, May 2017.

\bibitem{DualGANs}
Z.~{Yi}, H.~{Zhang}, P.~{Tan}, and M.~{Gong},
\newblock ``{DualGAN: Unsupervised Dual Learning for Image-to-Image
  Translation},''
\newblock {\em ArXiv:1704.02510}, Apr. 2017.

\bibitem{van2013wu}
David~C Van~Essen, Stephen~M Smith, Deanna~M Barch, Timothy~EJ Behrens, Essa
  Yacoub, Kamil Ugurbil, Wu-Minn~HCP Consortium, et~al.,
\newblock ``The wu-minn human connectome project: an overview,''
\newblock {\em Neuroimage}, vol. 80, pp. 62--79, 2013.

\bibitem{glasser2013}
Matthew~F Glasser, Stamatios~N Sotiropoulos, J~Anthony Wilson, Timothy~S
  Coalson, Bruce Fischl, Jesper~L Andersson, Junqian Xu, Saad Jbabdi, Matthew
  Webster, Jonathan~R Polimeni, et~al.,
\newblock ``{The minimal preprocessing pipelines for the Human Connectome
  Project},''
\newblock {\em Neuroimage}, vol. 80, pp. 105--124, 2013.

\bibitem{Yu2018}
Biting Yu, Luping Zhou, Lei Wang, Jurgen Fripp, and Pierrick Bourgerat,
\newblock ``{3D cGAN Based Cross-Modality MR Image Synthesis for Brain Tumor
  Segmentation},''
\newblock in {\em {International Symposium on Biomedical Imaging (ISBI)}},
  2018, pp. 626--630.

\end{thebibliography}

\end{document}